# Task Decomposition and Synchronization for Semantic Biomedical Image Segmentation

*Xuhua Ren, Lichi Zhang, Sahar Ahmad, Dong Nie, Fan Yang, Lei Xiang, Qian Wang\*, Member, IEEE
and Dinggang Shen\*, Fellow, IEEE*

*Abstract*— Semantic segmentation is essentially important to biomedical image analysis. Many recent works mainly focus on integrating the Fully Convolutional Network (FCN) architecture with sophisticated convolution implementation and deep supervision. In this paper, we propose to decompose the single segmentation task into three subsequent sub-tasks, including (1) pixel-wise image segmentation, (2) prediction of the class labels of the objects within the image, and (3) classification of the scene the image belonging to. While these three sub-tasks are trained to optimize their individual loss functions of different perceptual levels, we propose to let them interact by the task-task context ensemble. Moreover, we propose a novel sync-regularization to penalize the deviation between the outputs of the pixel-wise segmentation and the class prediction tasks. These effective regularizations help FCN utilize context information comprehensively and attain accurate semantic segmentation, even though the number of the images for training may be limited in many biomedical applications. We have successfully applied our framework to three diverse 2D/3D medical image datasets, including Robotic Scene Segmentation Challenge 18 (ROBOT18), Brain Tumor Segmentation Challenge 18 (BRATS18), and Retinal Fundus Glaucoma Challenge (REFUGE18). We have achieved top-tier performance in all three challenges.

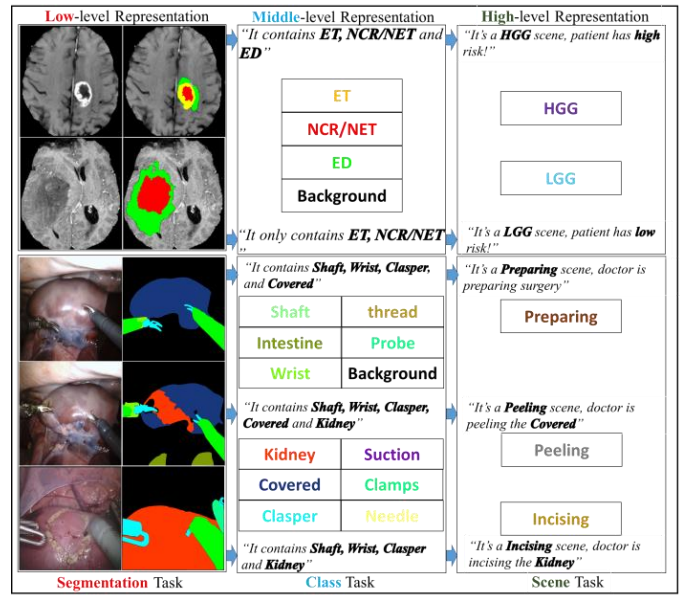

**Fig. 1**. Multi-level representations observed in BRATS18 and ROBOT18 datasets for semantic segmentation. Besides assigning pixel-wise label in the segmentation task, there are 4 classes and 2 scenes for BRATS18, 12 classes and 3 scenes for ROBOT18, respectively. The three tasks, as well as their multi-level representations, are closely coupled.

## I. INTRODUCTION

Semantic segmentation is a classical problem in the field of computer vision, where a pre-defined class label needs to be assigned to each pixel. The input image is thus divided into the regions corresponding to different class labels of a certain scene [1]. An optimal solution to segmentation usually relies on complicated representations including object class, location, scene and context.

Currently, most state-of-the-art semantic segmentation approaches are based on the FCN framework [2-7]. FCN has a powerful encoder to extract image features. Then, the decoder gradually fuses the high-level features at top layers of the encoder with the low-level features at bottom layers, which is essential for the decoder to generate high-quality semantic segmentation result [8].

The recent success of FCN can be attributed to the very deep network architecture, which pools the features into pyramid representations effectively. The deep supervision also contributes to the search for the network parameters. For example, EncNet [3] has a ResNet-101 encoder [9], and adopts

\* Corresponding authors: Qian Wang (wang.qian@sjtu.edu.cn), Dinggang Shen (dgshen@med.unc.edu)

Author, Xuhua Ren, Lichi Zhang, Xiang Lei and Qian Wang are with the Institute for Medical Imaging Technology, School of Biomedical Engineering, Shanghai Jiao Tong University, 200030, Shanghai, China.

Sahar Ahmad, Fan Yang, Dong Nie and Dinggang Shen are with the Department of Radiology and BRIC, University of North Carolina at Chapel Hill, Chapel Hill, North Carolina 27599, USA. Dinggang Shen is also with Department of Brain and Cognitive Engineering, Korea University, Seoul 02841, Republic of Korea.
.

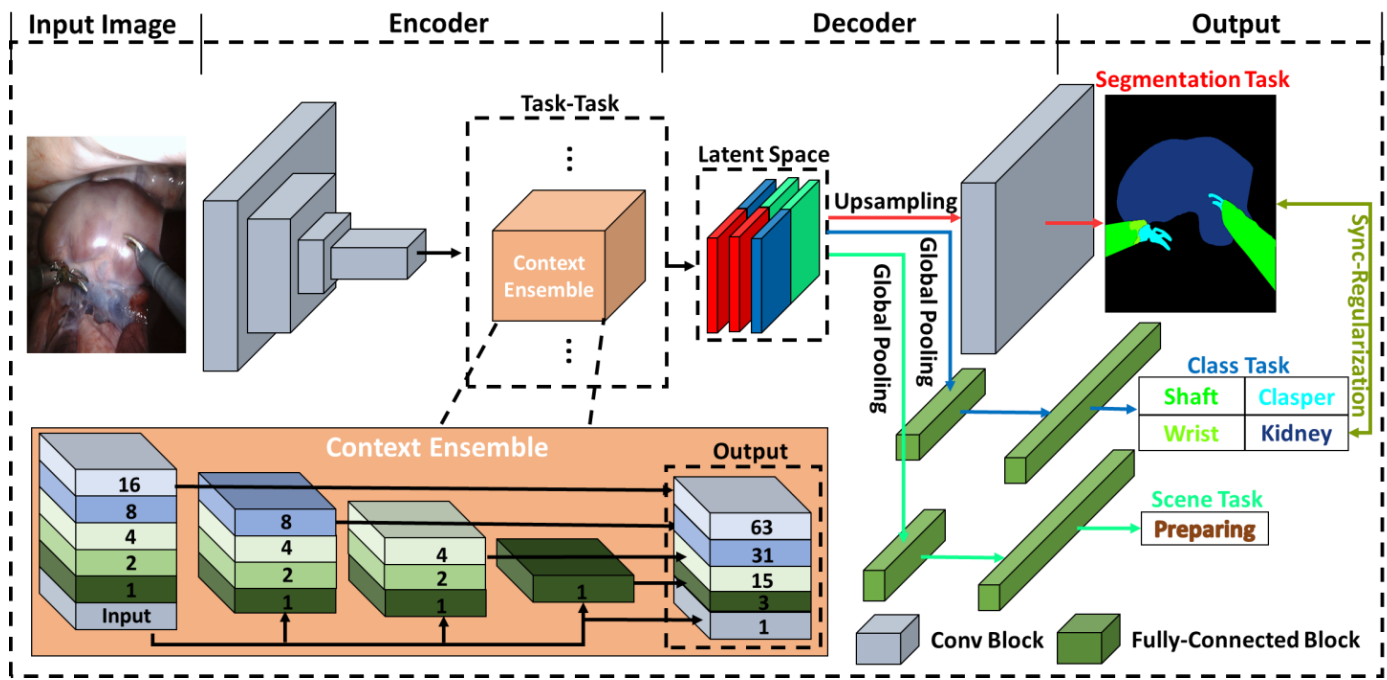

**Fig. 2.** The example of ROBOT18 with the proposed task decomposition framework for segmentation. The image is processed through the encoder and task-task context ensemble to arrive at the latent space. Then, the segmentation, class, and scene tasks are solved through individual decoders. A strong sync-regularization between the segmentation and class tasks is further used to augment the coherence of multi-task learning.

dilated convolution [10] in both encoder and decoder. A context encoding module strengthens deep supervision by incorporating semantic encoding loss. Pyramid scene parsing network (PSPNet) [2] has the traditional dilated FCN architecture for pixel prediction, while this network extends the pixel-level features to the specially designed global pyramid pooling features. The local and global cues together make the final prediction more reliable.

However, a sophisticated network in deep learning often needs a massive amount of data to train. Whereas a common yet critical challenge in biomedical image segmentation arises due to the limited size of the dataset – ROBOT18[1], REFUGE18[2] and BRATS18 [11], which are all widely used benchmark datasets in biomedical image segmentation and considered in this paper as well, have only 2,235, 400 and 285 subjects for training, respectively. To this end, it is essential to probe *how to adapt the network for small medical image datasets.*

While the datasets used in this paper are highly diverse, they all share the same goal of semantic segmentation.
1) Taking ROBOT18 for example (c.f. **Fig. 1**), the dataset is often used to validate instrument tracking and segmentation, which enables surgeons to conduct robot-assisted minimal invasive surgery (RMIS) more precisely. The image can be classified into three *scenes*, i.e., "preparing" surgery, "peeling" the covered kidney, and "incising" the kidney parenchyma. The existence of a certain *class* in the image can also be determined, followed by careful pixel-wise *segmentation*.
2) As a second example, BRATS18 is often used to validate automatic brain tumor segmentation, which is challenging due to the diversity of tumor location, shape, size and appearance [12]. Given a patient image, a clinician may first classify the *scene* as high-grade glioma (HGG) or low-grade glioma (LGG), since the two types of tumors usually appear different in images. Then, it can be determined whether the image renders certain sub-region of the tumor (or *class*, including ET, NCR/NET, ED, etc.), followed by careful pixel-wise *segmentation*.
3) REFUGE18 is used to evaluate and compare automated algorithms for glaucoma detection and optic disc/cup segmentation on retinal fundus images. Given a patient image, a clinician may first classify the *scene* as glaucoma patient or normal, as the patients may be different in the ratio between optic cup and disk. Then, it can be determined whether the image contains the *classes* of optic cup and disk, while the image can be processed through careful pixel-wise *segmentation* finally.

In general, it is evident that human expert segmentation is accurately conducted only if considering multi-level tasks and representations jointly.

Motivated by the above, we adopt task decomposition as a generalized solution to biomedical image segmentation (**first contribution**). Task decomposition is to perform multi-level tasks and representations which decompose a single task into several relative sub-tasks. While multi-level representations are essentially important to semantic segmentation, we decompose the segmentation task to three sub-tasks, i.e., to determine (1) pixel-wise segmentation, (2) object class, and (3) image scene. Although the three sub-tasks are closely coupled, traditional

[1] https://endovis.grand-challenge.org/
[2] https://refuge.grand-challenge.org/Home/

FCNs are only supervised by the pixel-wise segmentation loss, which is insufficient to decode the low-level task when ignoring high-level task/representation. By combing multi-level tasks, the network is able to comprehensively encode the context information for the low-level segmentation task.

The **second contribution** is the synchronization across different sub-tasks. Concerning the incoherence to learn multiple sub-tasks, we use task-task context ensemble to derive the common latent space, from which the features maps are forward to solve all sub-tasks jointly [13, 14]. Moreover, we propose a strong sync-regularization between the segmentation and class tasks, as the two tasks are very closely related with each other. Intuitively, in ROBOT18, if a class label (e.g., "kidney") is determined to have a certain 2D image, the pixel-wise segmentation should be consistent with some pixels labeled as "kidney". If inconsistent outcomes are detected, one may immediately realize the failure of the joint learning of the two sub-tasks. Therefore, the sync-regularization can supervise the network to better generalize multi-level representations.

The **third contribution** of our work is to implement semantic segmentation to diverse 2D/3D medical scenarios. Specifically, we adopt a shallow CNN as the encoder, followed by a spatial pyramid dilation module to ensemble context information for scene/class interpretation [15, 16]. Instead of using a Unet-like network [17], we adopt the PSPnet architecture [2]. We achieve top-tier performance in BRATS18, REFUGE18 and ROBOT18 challenges.

In general, we summarize our major contributions in this paper as follows:
1) We propose the task decomposition strategy to ease the challenging segmentation task in biomedical images.
2) We propose sync-regularization to coordinate the decomposed tasks, which gains advantage from multi-task learning toward image segmentation.
3) We build a practical framework for diverse biomedical image semantic segmentation and successfully apply it to three different challenge datasets.

The rest of this paper are organized as follows. We begin by reviewing literature reports related to semantic segmentation in Section II. The details of our framework and its components are presented in Section III. To verify the effectiveness of our method, extensive experiments are conducted and compared in Section IV. We conclude this work in Section V with extensive discussions in Section VI.

## II. RELATED WORKS

The FCN and its variants have demonstrated significant power on semantic segmentation. There are also many works to highlight the role of context information for the segmentation task.

**Semantic segmentation**: CNN and FCN have become state-of-the-art methods in semantic segmentation [18-22]. In FCN, fully connected layers are implemented as convolutions with large receptive fields, and segmentation is achieved using coarse class score maps obtained by feedforwarding the input image. The FCN network demonstrates impressive performance. Badrinarayanan et al. [23] presented SegNet which is the first encoder-decoder architecture for semantic pixel-wise segmentation, which consists of an encoder network and a corresponding decoder network followed by a pixel-wise classification layer. In semantic segmentation for medical images, Ronneberger et al. [6] presented U-net and proposed a training strategy with effective data augmentation for the limited number of annotated samples. Following the tremendous performance of the U-net architecture, Milletari et al. [24] proposed V-net for fully 3D image segmentation. Their CNN is trained end-to-end on MRI, and learns to segment the whole volume at once. There are also several attempts to address the degraded feature maps due to pooling, including Atrous Spatial Pyramid Pooling (ASPP), encoder-decoder, and dilated convolution [25]. Recently, some models such as PSPnet [2] and DeepLabV3+ [25] perform ASPP at various scales or apply dilated convolution in parallel. These models have been shown promising capability of handling semantic segmentation. In general, a typical segmentation network nowadays usually integrates: (1) the encoder module that gradually increases the receptive field to capture high-level context, (2) the decoder module that gradually recovers spatial information with skip connection, and (3) the dilated convolution that is effective to context ensemble. However, such a sophisticated network often needs large amount of data to optimize, which is a critical challenge in biomedical image segmentation task.

**Deep multi-task learning:** Deep multi-task learning aims at improving generalization capability by leveraging different domain-specific information [26]. There are many works of multi-task learning toward image segmentation and detection. Dai et al. [27] proposed Multi-task Network Cascades for instance-aware semantic segmentation, which is designed as cascaded structure to share the convolutional features. He et al. [28] proposed Mask R-CNN which jointly optimizes three tasks, i.e., detection, segmentation and classification, and outperformed single-task competitors. This approach efficiently detected objects in an image while simultaneously generating a high-quality segmentation mask for each instance. In the field of medical image multi-task learning, Ravi et al. [29] proposed a multi-task transfer learning DCNN with the aim of translating the 'knowledge' learned from non-medical images to medical segmentation tasks by simultaneously learning auxiliary tasks. Chen et al. [30] proposed a network which contains multi-level context features from the hierarchical architecture and explores auxiliary supervision for accurate gland segmentation. When incorporated with multi-task regularization during the training, the discriminative capability of latent features can be further improved. However, none of the methods mentioned above has combined multi-task learning with the idea of decomposing the segmentation task into sub-tasks and synchronizing them for better segmentation.

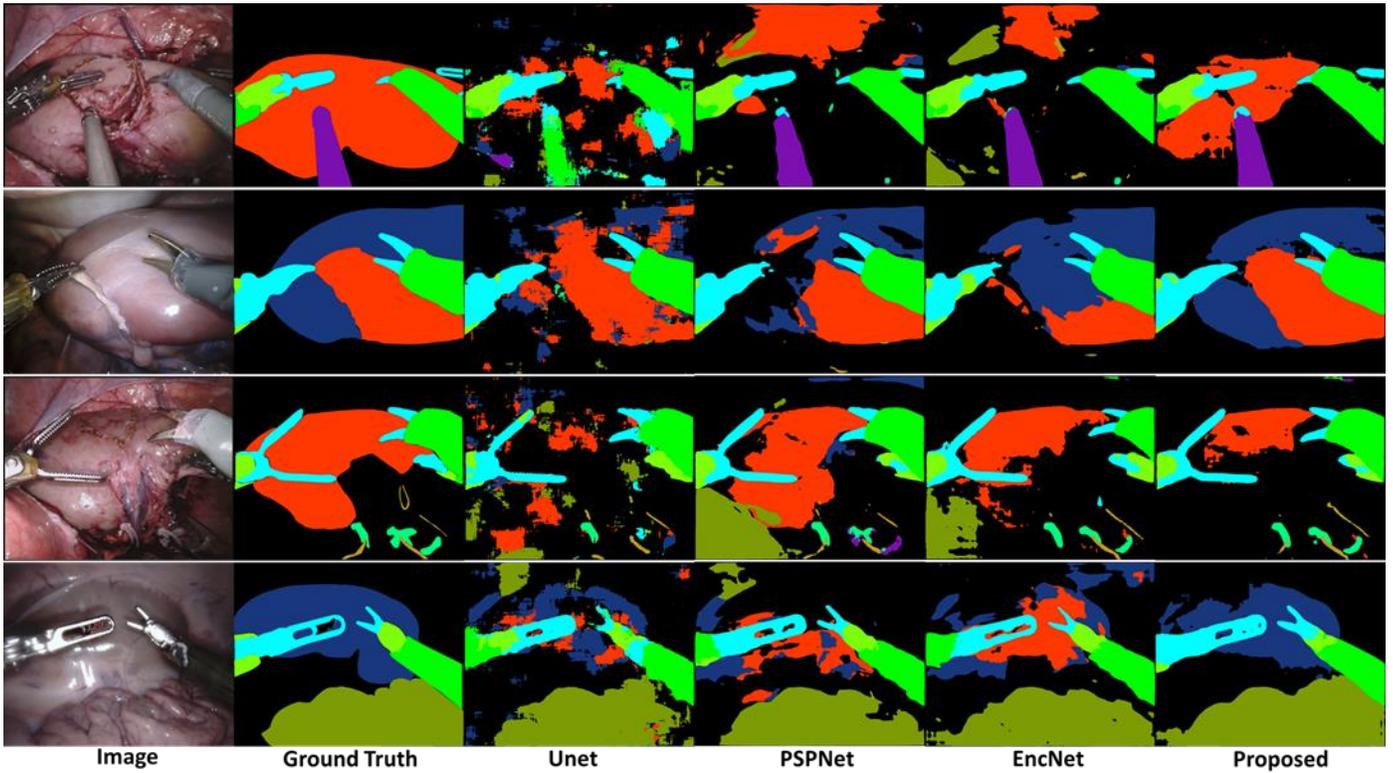

**Fig. 3**. Typical segmentation results for ROBOT18 by using different methods. From left to right are the original image, ground truth, Unet result, PSPNet result, EncNet result, and the proposed method's result.

Using deep multi-task learning with decomposition and sync-regularization could increase the generalization capability of the FCN model, and thus reduce the difficulty in semantic segmentation [3].

### III. PROPOSED METHOD

We propose the task decomposition framework as in **Fig. 2**. Given an input image (2D or 3D), we first use a dense convolutional encoder to extract feature maps. Because of the diversity of the input 2D/3D data, we design specific encoder for each of the three challenges in this paper. Then, we feed the extracted features to the task-task context ensemble module. The context ensemble module contains multi-scale dilated convolution, so the receptive fields are enlarged along the paths to combine features of different scales by different dilated rates. Moreover, the three parallel context ensemble modules are generated as task-task context ensemble module and each of module are connected by two branches which we called latent space. Finally, the network is decomposed from the latent space into three branches, corresponding to (1) the *segmentation* task, (2) the *class* task, and (3) the *scene* task. The decoders are trained for each decomposed task, including up sampling for the segmentation task, and also global average pooling and fully-connected layers for the class/scene tasks. Note that the three decomposed tasks share the same latent space for decoding.

#### A. Task Decomposition and Sync-Regularization

We aim to exploit multi-level representations for semantic segmentation. While state-of-the-art networks for segmentation often require tons of training data, it is possible to rely on the mutual dependency of the decomposed tasks to ease the parameter searching in FCN, especially concerning the limited numbers of medical images for training. We particularly decompose the semantic segmentation task into (1) pixel-wise image segmentation, (2) prediction of the object classes within the image, and (3) classification of the scene the image belonging to.

The three tasks are designed to optimize their individual loss functions. We utilize a hybrid loss $L_{seg}$ to supervise the low-level segmentation task:

$$L_{seg} = \sum_{x \in \Omega} CE(Y(x), \hat{Y}(x)) + (1 - IoU(Y(x), \hat{Y}(x))) \quad (1)$$

$CE$ is the cross-entropy between the segmentation ground truth $Y$ and the estimated $\hat{Y}$ for the image $x$ in the training set $\Omega$. $IoU$ divides the intersection area over the union between $Y$ and $\hat{Y}$. For both the high-level class/scene tasks, we adopt the same loss design as:

$$\begin{aligned} L_{cla} &= \sum_{x \in \Omega} BCE(C(x), \hat{C}(x)), \\ L_{scene} &= \sum_{x \in \Omega} BCE(S(x), \hat{S}(x)). \end{aligned} \quad (2)$$

$BCE$ is the binary cross-entropy, $C$ and $\hat{C}$ are the ground truth and the estimated class labels, respectively, and $S$ and $\hat{S}$ are the labels for the scene task.

## B. Task-Task Context Ensemble

The three tasks are trained to share the same latent space, while the task-task interaction is attained by context ensemble in our framework. The size of the receptive field, which is critical to explore context information, can often be too low [31]. To this end, we are inspired by Dlinknet [16] and cascade the dilated convolution for the context ensemble module as in in **Fig. 2**. Specifically, by coupling each pair of the decomposed tasks, the receptive fields enlarge their sizes, through the stacked dilated convolution with the dilated rates 63, 31, 15, 3, and 1, respectively, in the output of the module. The outputs of the three task-task CE modules are further paralleled and cascaded, which are resulted in the latent space. In detail, the three parallel context ensemble modules are generated and each of module are connected by two branches which we called this structure is latent space. Note that our context ensemble module can keep the spatial resolution/size of the feature maps, which significantly benefits segmentation accuracy.

## C. Sync-Regularization

As the network is decomposed into multiple branches for decoding, it is necessary to balance different tasks when generalizing the latent space. In particular, the class task can be regarded as a projection of the segmentation task. For example, the class task can determine whether "kidney" exists in a ROBOT18 image. One may also infer from the output of the segmentation task, and identify whether certain pixels are wrongly labeled as "kidney". By examining the deviation between the two tasks, we propose a novel sync-regularization strategy which has demonstrated powerful capability of learning the latent space jointly for better segmentation performance.

Specifically, we estimate the volume of each segmented label from the segmentation task's output, which is then converted to a boolean value (the pixel number of the specific class in segmentation map greater than 0 is set to 1). The vector of the boolean values, $P$, is then compared with the output of the class task, which results in the following loss function:

$$L_{sync} = \sum_{x \in \Omega} BCE(P(\hat{Y}(x)), \hat{C}(x)). \quad (3)$$

Here, $BCE$ is the binary cross-entropy. The sync-regularization provides feedbacks in back-propagation through the branch for the class task, and thus enforces consistency check to optimize the representations in the latent space. Note that the scene task is excluded from sync-regularization, as its association with the segmentation task is relatively loose. It's difficult to decompose segmentation map or classification vector into scene vector. E.g., converting segmentation maps of brain tumor to HGG/LGG classes is hard to implement.

In general, the total loss $L$ in our framework is

$$L = L_{seg} + w_1 \cdot L_{cla} + w_2 \cdot L_{scene} + w_3 \cdot L_{sync}, \quad (4)$$

where $w_1$, $w_2$, and $w_3$ are weights for tasks $L_{cla}$, $L_{scene}$, and $L_{sync}$, and also the hyper-parameters tuned in the experiments.

We tuned the $w$ parameters in sequence, and the set of $w$ value is 0.0, 0.2, 0.4, 0.8, 1.0.

## D. Encoder and Decoder

We design different encoders for 2D and 3D scenarios. In ROBOT18 and REFUGE18 that are 2D image datasets, we use VGG16 [32] pre-trained on ImageNet [33] as the encoders following LinkNet [34] setting. While for BRATS18 of 3D images, we start from Wang et al. [15] by using multiple layers of anisotropic and dilated convolution filters for encoder and decoder architectures.

Many medical image segmentation methods prefer to use the encoder-decoder structure and argue that it provides detailed boundary information to the object under consideration [35]. Here we have found that dilated convolution as in the PSPnet-like architecture can also extract multi-scale dense feature maps effectively and can achieve high performance in image segmentation. It is thus noted that connecting low-level and high-level layers with skip connection will possibly increase the difficulty of optimization in biomedical image segmentation. Particularly, for the decoders to the class/scene tasks, we first feed the feature maps through global average pooling [36], and then a fully connected layer decodes the desired outputs (c.f. **Fig. 2**).

## E. Training Procedure

It is known that the task-task interaction incurs additional difficulty to train in multi-task learning. To this end, we implement the task decomposition framework in a three-step training procedure:
1) We train the segmentation and class tasks together.
2) We utilize the early estimated parameters as initialization and refine the segmentation/class tasks with sync-regularization enforced.
3) We add in the scene task and refine all loss functions jointly.

Consequently, we obtain the fine-tuned FCN model that produces superior performance in the provided medical image datasets, while our training procedure tends to be robust in all experiments.

## IV. EXPERIMENTAL RESULTS

In this section, we first provide implementation details for our task decomposition framework. Then, we evaluate it on three diverse challenge datasets, including ROBOT18, BRATS18, and RUFUGE18.

### A. Implementation Details

Our implementation is based on Pytorch [37]. Regarding the hyper-parameters, the basic learning rate is set to 0.0001. For multi-task learning, the learning rate decreases gradually, i.e., $1 \times 10^{-4}$, $5 \times 10^{-5}$, $2 \times 10^{-5}$ after every 50 iterations. The momentum and weight decay are set to 0.9 and 0.0001, respectively.

In ROBOT18, we apply vertical/horizontal flip, RGB shift, random brightness (limit: 0.9-1.1) and random contrast (limit: 0.9-1.1) to augment the training data. For the validation dataset, no augmentation is adopted. Note that, for ROBOT18, there are 12 classes (i.e., shaft, intestine, wrist, thread, probe, kidney, covered, clasper, suction, clamps, needle and background) and

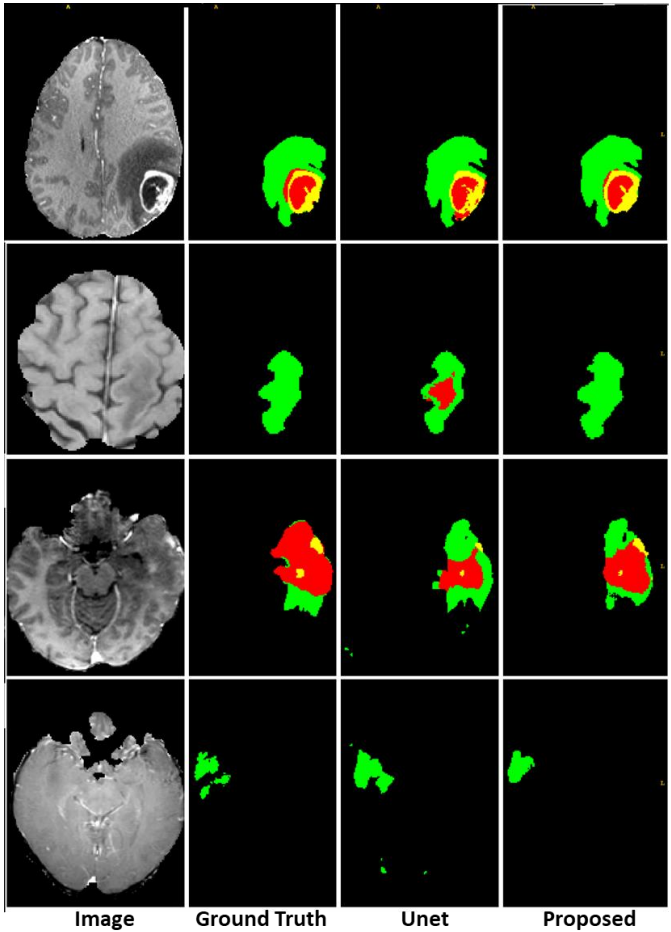

**Fig. 4**. Typical segmentation results for BRATS18 by using different methods. From left to right are for the original image, the ground truth, and results by Unet and our proposed method, respectively.

3 scenes (i.e., preparing, peeling and incising). We use 256 neurons in the hidden layers for both class and scene tasks (c.f. **Fig. 2**).

The second dataset BRATS18 is commonly used to validate brain tumor segmentation. For 3D medical images, the receptive field, model complexity and memory consumption of the network should be balanced. As a trade-off, we adopt the anisotropic setting that sets large receptive field in 2D slice but relatively small in the direction perpendicular to the slice. All images are thus cropped to the size of 144×144×19. For data augmentation, we only utilize random flip in three directions during training. There are 4 classes (i.e., ET, NCR/NET, ED and background) and 2 scenes (i.e., low-grade and high-grade gliomas) that are identified for BRATS18. We also use 128 neurons in the hidden layers, which is the same with ROBOT18.

The third dataset REFUGE18 is used to validate automated segmentation of optic disc/cup. We utilize same configuration of learning rate, data augmentation and training strategies with ROBOT18. There are 3 classes (i.e., optic disk, optic cup and background) and 2 scenes (i.e., glaucoma, normal). There are also 256 neurons used in the hidden layers.

### B. ROBOT: Robotic Scene Segmentation Challenge

There are 85 teams participating into the ROBOT18 challenge.

**Table 1**. Comparisons of single/multi-task learning, as well as the components of the proposed method, in solving ROBOT18 segmentation challenge. ("*Base+Class+Syn+Scene*" indicates the proposed method.)

|  | IoU: % | Dice: % |
|---|---|---|
| Single Task of *Segmentation* Only | | |
| ResNet34+Unet | 74.28 | 78.31 |
| ResNet50+Unet | 51.33 | 55.27 |
| VGG16+Unet | 75.82 | 79.17 |
| VGG16+CE+Unet | 75.90 | 79.23 |
| VGG16+CE+ASPP (*Base*) | **77.81** | **81.40** |
| Multiple Tasks of *Segmentation*, *Class*, and *Scene* | | |
| *Base+Class* (using Training Step 1) | 80.03 | 83.73 |
| *Base+Class+Sync* (Steps 1-2) | 81.99 | 85.56 |
| *Base+Class+Sync+Scene* (Steps 1-3) | **82.08** | **85.70** |

**Table 2**. Comparisons of the proposed method and other state-of-the-art methods in solving ROBOT18 segmentation challenge.

|  | IoU: % | Dice: % |
|---|---|---|
| PSPnet+ResNet34 | 69.86 | 73.06 |
| PSPnet+ResNet50 | 65.59 | 68.87 |
| Encnet+ResNet34 | 67.85 | 71.07 |
| Encnet+ResNet50 | 67.02 | 70.09 |
| Proposed (*Base+Class+Sync+Scene*) | **82.08** | **85.70** |

The challenge data has 2,235 training images, where the occurrence of certain class can be low. At the end of the challenge, 1,000 unseen images are released for test. The challenge is ranked on the mean IoU metric, which is computed per class and then averaged for the score of the entire test image. We can also compute Dice coefficient (Dice), a commonly used metric, to quantitatively assess the segmentation results.

To validate the design of our network, we compare several different settings and report the results in **Table 1**. First, we adopts VGG16 as the encoder while ASPP as the decoder ("*VGG16+CE+ASPP*"), while one may also choose other network architectures. The IoU/Dice scores of our implementation are 77.81% and 81.40%, both of which are higher than the alternatives. Particularly, we notice that the context ensemble (designated as "CE" in the table) module is effective even though only the segmentation task is considered. It is worth noting that, although the module is designed to couple a pair of decomposed tasks, we can also apply it to the case when only the single segmentation task is considered.

Second, with the network architecture ("*Base*") validated in the single-task learning, we further verify the contribution of the proposed task decomposition framework. Since there are three steps in our training procedure, we gradually add in the class/scene tasks and enforce the sync-regularization step by step. The experimental results in the bottom part of **Table 1** show that, with task decomposition, the multi-task solution to semantic segmentation outperforms the single-task solution, implying the effectiveness of the decomposed class/scene tasks toward image segmentation. Moreover, the sync-regularization also yields significant improvement for both IoU and Dice. Therefore, we conclude that the proposed task decomposition and sync-regularization are effective to the segmentation task (IoU: 82.08%; Dice: 85.70%).

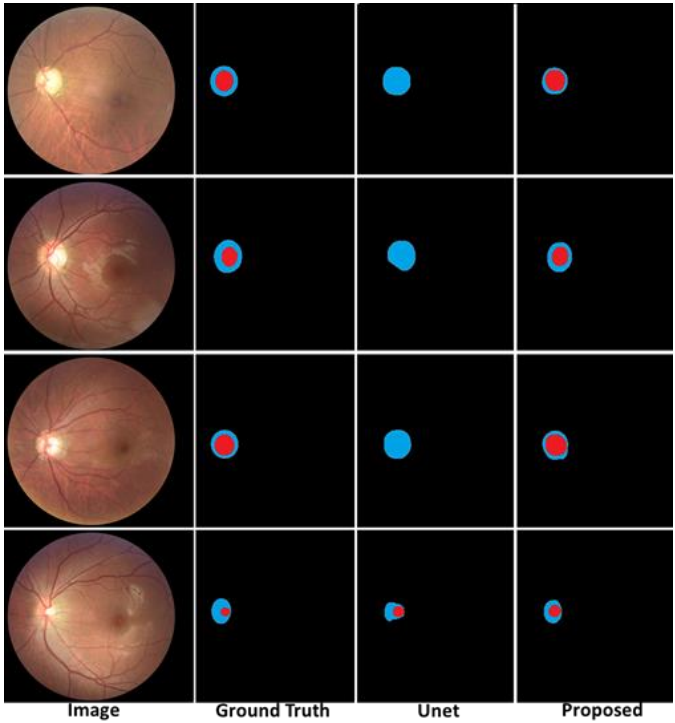

**Fig. 5**. Visual comparison on REFUGE18. Our proposed network achieves more accurate and detailed results. From left to right is original image, ground truth, and results by Unet and our proposed method, respectively.

Finally, we compare our proposed method with other state-of-the-art algorithms in **Table 2**. Considering the small size of the dataset, we adopt light-weighted encoders for PSPnet and Encnet. The results show that the proposed method ("*Base+Class+Syn+Scene*") outperforms all methods under comparison in the validation set.

Moreover, our method has demonstrated top-tier performance in the on-site testing set in ROBOT18 (rank second, IoU=61%, compared to 62% of the challenge winner). Note that the on-site test data IoU score is different with what is shown in **Table 2**, since the images in the on-site testing set and the validation set are not coherent with each other. We have also provided visual inspection of typical segmentation results of ROBOT18 in **Fig. 3**, where our method clearly performs better than the alternatives under consideration.

*C. BRATS: Brain Tumor Segmentation Challenge*

BRATS is focusing on the evaluation of state-of-the-art methods for the segmentation of brain tumors in multi-modal magnetic resonance (MR) scans [38]. There are more than 200 teams participating in the BRATS18. In our work there are 285 subjects used for training, each of which comes with a 240×240×155 sized 3D image. We also use the data of 66 subjects for on-site validation. Slightly different from ROBOT18, the BRATS18 challenge uses Dice and Hausdorff distance (HD) as the performance metrics. All comparisons are conducted on the on-site validation data, while the results are submitted directly through the official website[3]. The organizers return the ranks of all participants, as our submission ranks first in BRATS18. In order to validate the design of our network, we compare several different settings and report respective results in **Table 3**, which is similar to early experiments in ROBOT18.

[3] https://www.med.upenn.edu/sbia/brats2018

**Table 3**. Comparisons of single/multi-task learning, as well as the components of the proposed method, in solving BRATS18 segmentation challenge. ("*Base+Class+Syn+Scene*" indicates the proposed method.)

|  | Dice: % | HD: mm |
|---|---|---|
| Single Task of *Segmentation* Only | | |
| Wnet+Unet | 79.67 | 17.09 |
| Wnet +CE+Unet | 79.69 | 16.43 |
| Wnet +CE+ASPP (*Base*) | **80.15** | **15.75** |
| Multiple Tasks of *Segmentation*, *Class*, and *Scene* | | |
| *Base+Class* (using Training Step 1) | 80.41 | 10.30 |
| *Base+Class+Sync* (Steps 1-2) | 80.87 | 10.14 |
| *Base+Class+Sync+Scene* (Steps 1-3) | **80.88** | **9.74** |

**Table 4**. Comparisons of the proposed method and other state-of-the-art methods in solving BRATS18 segmentation challenge.

|  | Dice: % | HD: mm |
|---|---|---|
| Unet | 79.45 | 12.76 |
| Deepmedic | 80.42 | 10.54 |
| GTNet | **82.29** | 11.30 |
| Proposed (*Base+Class+Sync+Scene*) | 80.88 | **9.74** |

**Table 5**. Comparisons of our result and other top-ranked results submitted to the on-site validation of BRATS18 segmentation challenge.[4]

| Team | Dice: % | | | HD: mm | | |
|---|---|---|---|---|---|---|
|  | ET | WT | TC | ET | WT | TC |
| NVDL | 82.5 | 91.2 | 87.0 | 4.0 | 4.5 | 6.8 |
| MIC | 80.9 | 91.3 | 86.3 | **2.4** | 4.3 | **6.5** |
| BIGS2 | 80.5 | 91.0 | 85.1 | 2.8 | 4.8 | 7.5 |
| Proposed | **83.2** | **91.5** | **88.3** | 2.9 | **3.9** | 7.7 |

Our implementation adopts state-of-the-art "*Wnet*" as the encoder for 3D medical image segmentation. Meanwhile, we use ASPP as the decoder. Therefore, we term our implementation as "*Wnet+CE+ASPP*". The Dice/HD scores of our method are 80.15% and 15.75 mm on the on-site validation set, both of which are better than other cases (c.f. **Table 3**, the *single-task* setting). Particularly, we notice that the context ensemble module is effective even though only the single segmentation task is considered.

We also conduct the experiment of sematic segmentation using the proposed task decomposition strategy, and compare its performance with the alternative configurations, including "*Base+Class*", "*Base+Class+Sync*", and "*Base+Class+Sync+Scene*" in **Table 3**. The findings in **Table 3** are also similar with those in **Table1**: using the whole three steps as the multi-task solution can further improve the segmentation performance; the proposed task decomposition and sync-regularization are also considered as beneficial to the segmentation task for the BRATS18 dataset (Dice: 80.88 %; HD: 9.74 mm). Besides, we compare our proposed method with other state-of-the-art algorithms in **Table 4**. For single state-of-the-art models like 3D Unet [39] and Deepmedic [40], our method could outperform these methods on both evaluation metrics. Comparing our method with the winner of the BRATS challenge in 2017 (GTNet) [15], we also have a better performance on HD score while worse Dice. Note that GTNet requires to train 9 different segmentation models for the coarse-to-fine ensemble.

[4] https://www.cbica.upenn.edu/BraTS18/

**Table 6**. Comparisons of single/multi-task learning, as well as the components of the proposed method, in solving REFUGE18 segmentation challenge. ("*Base+Class+Syn+Scene*" indicates the proposed method.)

| | IoU(%) | Dice(%) |
|---|---|---|
| Single Task of *Segmentation* Only | | |
| ResNet34+Unet | 75.04 | 84.78 |
| ResNet50+Unet | 76.85 | 85.98 |
| VGG16+Unet | 81.26 | 89.13 |
| VGG16+CE+Unet | 82.47 | 90.43 |
| VGG16+CE+ASPP (*Base*) | **82.67** | **90.72** |
| Multiple Tasks of *Segmentation*, *Class*, and *Scene* | | |
| *Base+Class* (using Training Step 1) | 82.91 | 91.02 |
| *Base+Class+Sync* (Steps 1-2) | 83.44 | 91.21 |
| *Base+Class+Sync+Scene* (Steps 1-3) | **83.49** | **91.29** |

**Table 7**. Comparisons of the proposed method and other state-of-the-art methods in solving REFUGE18 segmentation challenge.

| | IoU: % | Dice: % |
|---|---|---|
| PSPnet+ResNet34 | 75.52 | 84.99 |
| PSPnet+ResNet50 | 72.96 | 82.62 |
| Encnet+ResNet34 | 75.57 | 85.31 |
| Encnet+ResNet50 | 73.02 | 83.46 |
| Proposed (*Base+Class+Sync+Scene*) | **83.49** | **91.29** |

Moreover, our method has obtained the top-ranked score on the on-site validation set in BRATS18 as in **Table 5**. Our final submission achieves 83.2, 91.5, 88.3 for Dice (%) and 2.9, 3.9, 7.7 for HD (mm) on three foreground classes (i.e., ET, WT, and TC). We have also provided visual inspection of the segmentation results of BRATS18 in **Fig. 4**.

*D. REFUGE: Retinal Fundus Glaucoma Challenge*

The number of teams participating in the REFUGE18 is more than 400. Here we use 400 training images in the REFUGE18 challenge dataset. At the end of the challenge, 400 unseen images are released for test. The challenge is ranked on the mean Dice metric, which is computed per class and then averaged for the score of the entire test image. We can also compute Intersection over Union (IoU) to quantitatively assess the segmentation results.

We also conduct the experiments of performance evaluation for the REFUGE18 dataset, and compare the results using several different settings in single-task learning first. **Table 6** verifies the IoU and the Dice measurements for the experimental results. While our implementation adopts VGG16 as the encoder and ASPP as the decoder ("*VGG16+CE+ASPP*"), the IoU/Dice scores of our method are 82.67% and 90.72%, both of which are higher than other choices.

The experimental results for the proposed task decomposition framework are also shown in **Table 6**, where we also evaluate the validity of the three training steps for the REFUGE18 dataset. Similar with what are previously shown in **Table 1** and **Table 3**, the segmentation performance using all three steps outperforms the cases when the class/scene tasks and the sync-regularization component are not fully applied. The IoU and the Dice scores for our method ("*Base+Class+Sync+Scene*") reach 83.49% and 91.29%, respectively.

Finally, we compare our proposed method with other state-of-the-art algorithms in **Table 7**. Here we adopt light-weighted encoders for PSPnet and Encnet due to the limited size of the training datasets. The results show that the proposed method outperforms all methods under comparison in the validation set. Moreover, the on-site score of our method for REFUGE18 can be mostly comparable with the winner score of the challenge (IoU=91%, compared to 92% of the winner). We have also provided visual inspection of the segmentation results of REFUGE18 in **Fig. 5**.

V. DISCUSSION AND CONCLUSION

As a summary, we have proposed an effective task decomposition framework for semantic segmentation of diverse biomedical images in this work. Specifically, we decompose the very challenging semantic segmentation task to seek for helps from auxiliary class/scene tasks. The decomposed tasks are associated with low-level to high-level representations and reduce the complexity to solve image segmentation. Moreover, in addition to context ensemble in the latent space, we propose sync-regularization to penalize the deviation between different tasks and to coordinate multi-task learning for the sake of semantic segmentation. We have conducted comprehensive experiments on three very diverse yet popular medical image datasets. Our results are currently top-ranking in all three challenges.

In this work, the entire training phase of our method takes about 4, 7, 2 hours on a single NVIDIA Titan X GPU for ROBOT18, BRATS18 and REFUGE18 dataset, respectively. It is worth noting that it costs only 0.1s (2s) to generate the final segmentation map for one 2D (3D) image in the testing stage. While the accuracy of the CNN for medical image segmentation is mostly comparable to (or better than) state-of-the-art algorithms, the runtime is shown to outperform other multi-task segmentation approach (e.g., Mask R-CNN [28]) for medical image segmentation. This will be particularly beneficial in large clinical environments where hundreds or sometimes thousands of people are screened every day.

There are some limitations for our proposed method. Our proposed encoder and decoder try to find a trade-off between depth and computational feasibility. While increasing depth addresses larger semantic regions during processing, contraction elements such as max-pooling reduces the number of parameters but loses information about specific locations. In contrast, dilated convolution operation may lead to increased parameters but keep more information in detail. Although the proposed method has shown its promising results in the experiments, it would be interesting how to obtain an optimal architecture which could keep necessary information but cost less in computation. Moreover, in this work our sync-regularization is a single-directional loss which only matches the segmentation task with classification one. However, the reversible loss is a more reasonable way to measure the variance between the two tasks considering classification task may misguide the optimization of segmentation task. Finally, we only explore the biomedical image datasets which only have limited number of images in our work. It would be valuable to explore this further and see how the method performs on more traditional natural images, particularly when varying the amount of training data.

In future work, we will further improve method from the following two aspects. 1) We will simplify the sync-regularization as the single destination loss in the network. We need to explore a feasible way to build a reversible loss between segmentation and classification tasks in our proposed network. 2) Our proposed innovation modules, including task decomposition and synchronization, can be further explored especially in a large amount of training dataset.

## VI. REFERENCE

[1] J. Long, E. Shelhamer, and T. Darrell, "Fully convolutional networks for semantic segmentation," in *Proceedings of the IEEE conference on computer vision and pattern recognition*, 2015, pp. 3431-3440.
[2] H. Zhao, J. Shi, X. Qi, X. Wang, and J. Jia, "Pyramid scene parsing network," in *IEEE Conf. on Computer Vision and Pattern Recognition (CVPR)*, 2017, pp. 2881-2890.
[3] H. Zhang *et al.*, "Context encoding for semantic segmentation," in *The IEEE Conference on Computer Vision and Pattern Recognition (CVPR)*, 2018.
[4] L.-C. Chen, G. Papandreou, I. Kokkinos, K. Murphy, and A. L. Yuille, "Deeplab: Semantic image segmentation with deep convolutional nets, atrous convolution, and fully connected crfs," *IEEE transactions on pattern analysis and machine intelligence,* vol. 40, no. 4, pp. 834-848, 2018.
[5] Z. Zhang, X. Zhang, C. Peng, D. Cheng, and J. Sun, "ExFuse: Enhancing Feature Fusion for Semantic Segmentation," *arXiv preprint arXiv:1804.03821,* 2018.
[6] O. Ronneberger, P. Fischer, and T. Brox, "U-net: Convolutional networks for biomedical image segmentation," in *International Conference on Medical image computing and computer-assisted intervention*, 2015, pp. 234-241: Springer.
[7] H. Noh, S. Hong, and B. Han, "Learning deconvolution network for semantic segmentation," in *Proceedings of the IEEE international conference on computer vision*, 2015, pp. 1520-1528.
[8] V. Badrinarayanan, A. Kendall, and R. Cipolla, "Segnet: A deep convolutional encoder-decoder architecture for image segmentation," *arXiv preprint arXiv:1511.00561,* 2015.
[9] K. He, X. Zhang, S. Ren, and J. Sun, "Deep residual learning for image recognition," in *Proceedings of the IEEE conference on computer vision and pattern recognition*, 2016, pp. 770-778.
[10] F. Yu and V. Koltun, "Multi-scale context aggregation by dilated convolutions," *arXiv preprint arXiv:1511.07122,* 2015.
[11] S. Bakas *et al.*, "Advancing the cancer genome atlas glioma MRI collections with expert segmentation labels and radiomic features," *Scientific data,* vol. 4, p. 170117, 2017.
[12] W. Xiong, C.-K. Chui, and S.-H. Ong, "Focus, Segment and Erase: An Efficient Network for Multi-Label Brain Tumor Segmentation," *Ratio,* vol. 5, no. 13.4, p. 16.8.
[13] S. Ruder, "An overview of multi-task learning in deep neural networks," *arXiv preprint arXiv:1706.05098,* 2017.
[14] Z. Chen, V. Badrinarayanan, C.-Y. Lee, and A. Rabinovich, "GradNorm: Gradient Normalization for Adaptive Loss Balancing in Deep Multitask Networks," *arXiv preprint arXiv:1711.02257,* 2017.
[15] G. Wang, W. Li, S. Ourselin, and T. Vercauteren, "Automatic brain tumor segmentation using cascaded anisotropic convolutional neural networks," in *International MICCAI Brainlesion Workshop*, 2017, pp. 178-190: Springer.
[16] L. Zhou, C. Zhang, and M. Wu, "D-LinkNet: LinkNet with Pretrained Encoder and Dilated Convolution for High Resolution Satellite Imagery Road Extraction," in *Proceedings of the IEEE Conference on Computer Vision and Pattern Recognition Workshops*, 2018, pp. 182-186.
[17] O. Ronneberger, P. Fischer, and T. Brox, "U-Net: Convolutional Networks for Biomedical Image Segmentation," in *medical image computing and computer assisted intervention*, 2015, pp. 234-241.
[18] D. Shen, G. Wu, and H.-I. Suk, "Deep learning in medical image analysis," *Annual review of biomedical engineering,* vol. 19, pp. 221-248, 2017.
[19] H. Chen, Q. Dou, L. Yu, J. Qin, and P.-A. Heng, "VoxResNet: Deep voxelwise residual networks for brain segmentation from 3D MR images," *NeuroImage,* 2017.
[20] A. V. Dalca, J. Guttag, and M. R. Sabuncu, "Anatomical Priors in Convolutional Networks for Unsupervised Biomedical Segmentation," in *Proceedings of the IEEE Conference on Computer Vision and Pattern Recognition*, 2018, pp. 9290-9299.
[21] X. Xu *et al.*, "Quantization of fully convolutional networks for accurate biomedical image segmentation," *Preprint at https://arxiv.org/abs/1803.04907,* 2018.
[22] K.-L. Tseng, Y.-L. Lin, W. Hsu, and C.-Y. Huang, "Joint sequence learning and cross-modality convolution for 3d biomedical segmentation," in *Computer Vision and Pattern Recognition (CVPR), 2017 IEEE Conference on*, 2017, pp. 3739-3746: IEEE.
[23] V. Badrinarayanan, A. Kendall, R. J. I. t. o. p. a. Cipolla, and m. intelligence, "Segnet: A deep convolutional encoder-decoder architecture for image segmentation," vol. 39, no. 12, pp. 2481-2495, 2017.
[24] F. Milletari, N. Navab, and S.-A. Ahmadi, "V-net: Fully convolutional neural networks for volumetric medical image segmentation," in *2016 Fourth International Conference on 3D Vision (3DV)*, 2016, pp. 565-571: IEEE.
[25] L.-C. Chen, Y. Zhu, G. Papandreou, F. Schroff, and H. Adam, "Encoder-decoder with atrous separable convolution for semantic image segmentation," *arXiv preprint arXiv:1802.02611,* 2018.
[26] Y. Zhang and Q. J. a. p. a. Yang, "A survey on multi-task learning," 2017.
[27] J. Dai, K. He, and J. Sun, "Instance-aware semantic segmentation via multi-task network cascades," in *Proceedings of the IEEE Conference on Computer Vision and Pattern Recognition*, 2016, pp. 3150-3158.
[28] K. He, G. Gkioxari, P. Dollár, and R. Girshick, "Mask r-cnn," in *Computer Vision (ICCV), 2017 IEEE International Conference on*, 2017, pp. 2980-2988: IEEE.
[29] C. Zu *et al.*, "Label-aligned multi-task feature learning for multimodal classification of Alzheimer's disease and mild cognitive impairment," vol. 10, no. 4, pp. 1148-1159, 2016.
[30] H. Chen, X. Qi, L. Yu, Q. Dou, J. Qin, and P.-A. J. M. i. a. Heng, "DCAN: Deep contour-aware networks for object instance segmentation from histology images," vol. 36, pp. 135-146, 2017.
[31] W. Luo, Y. Li, R. Urtasun, and R. Zemel, "Understanding the effective receptive field in deep convolutional neural networks," in *Advances in neural information processing systems*, 2016, pp. 4898-4906.
[32] C. Szegedy *et al.*, "Going deeper with convolutions," 2015: Cvpr.
[33] J. Deng, W. Dong, R. Socher, L.-J. Li, K. Li, and L. Fei-Fei, "Imagenet: A large-scale hierarchical image database," in *Computer Vision and Pattern Recognition, 2009. CVPR 2009. IEEE Conference on*, 2009, pp. 248-255: IEEE.
[34] A. Chaurasia and E. Culurciello, "Linknet: Exploiting encoder representations for efficient semantic segmentation," in *2017 IEEE Visual Communications and Image Processing (VCIP)*, 2017, pp. 1-4: IEEE.
[35] G. Lin, A. Milan, C. Shen, and I. D. Reid, "RefineNet: Multi-path Refinement Networks for High-Resolution Semantic Segmentation," in *Cvpr*, 2017, vol. 1, no. 2, p. 5.
[36] M. Lin, Q. Chen, and S. Yan, "Network in network," *arXiv preprint arXiv:1312.4400,* 2013.
[37] A. Paszke *et al.*, "Automatic differentiation in pytorch," 2017.
[38] B. H. Menze *et al.*, "The multimodal brain tumor image segmentation benchmark (BRATS)," *IEEE transactions on medical imaging,* vol. 34, no. 10, p. 1993, 2015.
[39] F. Milletari, N. Navab, and S.-A. Ahmadi, "V-net: Fully convolutional neural networks for volumetric medical image segmentation," in *3D Vision (3DV), 2016 Fourth International Conference on*, 2016, pp. 565-571: IEEE.

[40] K. Kamnitsas *et al.*, "DeepMedic for brain tumor segmentation," in *International Workshop on Brainlesion: Glioma, Multiple Sclerosis, Stroke and Traumatic Brain Injuries*, 2016, pp. 138-149: Springer.